\documentclass[conference]{IEEEtran}
\IEEEoverridecommandlockouts
\PassOptionsToPackage{bookmarks={false}}{hyperref}
\usepackage{cite}
\usepackage[inline]{enumitem}
\usepackage{amsmath,amssymb,amsfonts}

\usepackage{subfig}
\usepackage{graphicx}
\usepackage{textcomp}
\usepackage{xcolor}
\usepackage{soul}
\usepackage{caption}
\usepackage{epstopdf}
\usepackage{array}
\usepackage{nicefrac}
\usepackage{siunitx}
\usepackage{adjustbox}

\usepackage[colorinlistoftodos]{todonotes}
\makeatletter
\usepackage[utf8]{inputenc}
\newcommand{\inlineitem}[1][]{%
\ifnum\enit@type=\tw@
    {\descriptionlabel{#1}}
  \hspace{\labelsep}%
\else
  \ifnum\enit@type=\z@
       \refstepcounter{\@listctr}\fi
    \quad\@itemlabel\hspace{\labelsep}%
\fi}
\makeatother
\def\BibTeX{{\rm B\kern-.05em{\sc i\kern-.025em b}\kern-.08em
    T\kern-.1667em\lower.7ex\hbox{E}\kern-.125emX}}

\newcommand{\comment}[1]{ }

\usepackage{adjustbox}

\usepackage{lipsum}

\usepackage{amsmath}
\usepackage{float}
\usepackage{soul}
\usepackage{booktabs}
\usepackage{tabularx}
\usepackage{color}
\usepackage[algo2e]{algorithm2e}
\usepackage{lipsum,multicol}
\usepackage{array,multirow}
\usepackage{lettrine}
\usepackage{gensymb}
\usepackage{mathtools, nccmath}

\usepackage{algorithm}
\usepackage[font=footnotesize, labelfont={bf, color=accessblue}, margin=.75cm, labelsep = period]{caption}
\usepackage{algorithmic}
\usepackage{lscape}
\graphicspath{{./}{./Figure/}}
\usepackage{enumerate}
\usepackage{epstopdf}
\usepackage{float}
\usepackage{url}
\usepackage{booktabs}
\usepackage{cite}
\usepackage{array}

\newcommand{\myitemizebeginless}{\begin{list}{$\bullet$}
{
 \setlength{\leftmargin}{0.09cm}
 \setlength{\parsep}{0.0cm}
 \setlength{\itemsep}{0.05cm}
 \setlength{\topsep}{0.0cm}
}}
\newcommand{\myitemizeendless}{\end{list}}

\usepackage{tabularx}
\def\BibTeX{{\rm B\kern-.05em{\sc i\kern-.025em b}\kern-.08em T\kern-.1667em\lower.7ex\hbox{E}\kern-.125emX}}

\begin{document}

\title{Deep-Learning-Based Device Fingerprinting for Increased LoRa-IoT Security: Sensitivity to Network Deployment Changes
\thanks{An IEEE-formatted version of this article is published in IEEE Network. Personal use of this material is permitted. Permission from IEEE must be obtained for all other uses, in any current or future media, including reprinting/republishing this material for advertising or promotional purposes, creating new collective works, for resale or redistribution to servers or lists, or reuse of any copyrighted component of this work in other works.}
}

\author{\IEEEauthorblockN{Bechir Hamdaoui and Abdurrahman Elmaghbub}
\IEEEauthorblockA{{School of Electrical Engineering and Computer Science}, {Oregon State University}, Corvallis, Oregon, USA\\
\{hamdaoui,elmaghba\}@oregonstate.edu} }

\maketitle

\begin{abstract}
Deep-learning-based device fingerprinting has recently been recognized as a key enabler for automated network access authentication.
Its robustness to impersonation attacks due to the inherent difficulty of replicating physical features is what distinguishes it from conventional cryptographic solutions.
Although device fingerprinting has shown promising performances, its sensitivity to changes in the network operating environment still poses a major limitation.
This paper presents an experimental framework that aims to study and overcome the sensitivity of LoRa-enabled device fingerprinting  to such changes.
We first begin by describing RF datasets we collected using our LoRa-enabled wireless device testbed. We then propose a new fingerprinting technique that exploits out-of-band distortion information caused by hardware impairments to increase the fingerprinting accuracy.
Finally, we experimentally study and analyze the sensitivity of LoRa RF fingerprinting to various network setting changes.
Our results show that fingerprinting does relatively well when the learning models are trained and tested under the same settings. However, when trained and tested under different settings, these models exhibit moderate sensitivity to channel condition changes and severe sensitivity to protocol configuration and receiver hardware changes when IQ data is used as input.
However, with FFT data is used as input, they perform poorly under any change.

\end{abstract}



\section{Introduction}
\label{sec:Introducation}
LoRa (short for Long Range) wireless technology has become the de facto physical platform for a wide range of Internet of Things (IoT) networks and applications. In recent years, hundreds of IoT application developers have adopted LoRaWAN protocol, a wide-area network (WAN) protocol built on top of the LoRa physical layer as the solution for enabling long-range connectivity between IoT devices.
Recent studies estimate that the global number of connected IoT devices, currently at 12.3 Billion, is projected to reach 27 Billion globally by 2025\mbox{~\cite{IoTAnalytics-21}}. With the increased adoption of LoRa technology coupled with the growing numbers of connected IoT devices, there is undoubtedly an urgent need for efficient security mechanisms to protect such emerging LoRa-enabled devices and networks.

Physical-layer security approaches that are based on wireless RF (radio frequency) fingerprints have recently emerged as potential device identification solutions for automated monitoring of authorized IoT network access\mbox{~\cite{abbas2021improving}}. What distinguishes these fingerprinting solutions from higher-layer ones is the inability of attackers to replicate the fingerprints. As a result, they are foreseen to complement to conventional cryptographic security approaches to further protect networks against unauthorized access.

RF fingerprinting consists of extracting features from the received RF signals and using them to uniquely identify the transmitting devices~\cite{jian2020deep}. However, the feature selection process has, until recently, been performed manually, thus requiring RF signal domain expertise and many trial-and-error iterations to find the best feature set.
Driven by the recent success of applying deep learning to image recognition and natural language processing~\cite{masone2021survey}, attempts in using deep learning to solve wireless networking problems~\cite{elsayed2019ai}, including device fingerprinting~\cite{hamdaoui2020deep}, have also yielded promising results.
Deep-learning-driven RF fingerprinting techniques can automatically process raw RF signals and extract the features without needing domain knowledge~\cite{cekic2020robust}. For instance, they can exploit the random distortions in the received RF signals that are caused by the device hardware impairments to extract signatures that are unique to the devices~\cite{sankhe2019oracle}.
However, as the radio technology continues to advance, the variation in the impairments across different devices also continues to decrease, thereby resulting in a decrease in the distances between devices in the feature space. This decrease makes the task of device fingerprinting more challenging, as it becomes more sensitive even to the slightest changes in the wireless channel condition and/or the network deployment setting.
For instance, in \cite{al2020exposing}, the authors attributed the sensitivity of deep learning models to channel overfitting, and 
concluded that equalizing I/Q (in-phase/quadrature) data can mitigate this problem.
An artificial noise adding approach was proposed in \cite{zhou2021robust} to increase the robustness of ZigBee device identification against channel variability.
%
In \cite{cekic2020wireless}, the authors studied the location sensitivity and proposed augmentation and estimation strategies to promote generalization across time and location. These studies are critical to the deployment of RF fingerprinting techniques in real-world. However, they were limited to WiFi, ADS-B (automatic dependent surveillance-broadcast) and ZigBee data and focused on changes along the time and location dimensions only.

In this paper, we experimentally study the sensitivity of LoRa device fingerprinting to such changes, assess its ability to classify different LoRa IoT devices with identical hardware components, and propose an approach to improve its identification accuracy.
More specifically, we:
\begin{itemize}
\item {\bf Released massive RF datasets} collected from $25$ LoRa devices that cover a number of indoor and outdoor scenarios, with varying deployment settings, including the experiment location and time, the LoRa protocol configuration, and the receiver hardware.

\item {\bf Presented a new technique} that exploits out-of-band distortion information caused by hardware impairments to provide unique, device-specific signatures that are then leveraged to increase the fingerprinting accuracy.

\item {\bf Conducted an experimental study} that discloses the sensitivity of deep-learning-based LoRa RF fingerprinting to various network deployment changes while considering two different data input representations of the sampled RF signals: time-domain IQ and frequency-domain FFT (fast Fourier transform).

\end{itemize}

Our results reveal that the fingerprinting techniques do relatively well when trained and tested under the same deployment settings, with FFT representation as input outperforming the IQ representation. However, if trained and tested under different settings, when IQ data is used as input to the learning models, these techniques show moderate sensitivity to channel condition variations and deep sensitivity to system setting variations such as changes in the LoRa protocol configuration and receiver hardware. However, when FFT data is used as input, they perform exceptionally poorly, regardless of the type of change.

The rest of the paper is organized as follows. Section \ref{sec:te}  describes the testbed and the collected datasets.
Section \ref{sec:hw} explains the hardware impairments and highlights their impact on the out-of-band (OOB) spectrum emissions, and Section~\ref{sec:proposed} presents the proposed technique leveraging OOB emissions to increase fingerprinting accuracy. Section \ref{sec:Evaluation} presents result findings showing the vulnerability of RF fingerprinting to various changes in the network operating environment. Section~\ref{open-challenges} discusses open challenges and future research directions. Lastly, Section \ref{sec:Conclusion} concludes the article.

\section{Testbed and Dataset Collection}
\label{sec:te}
\subsection{LoRa Device Experimental Testbed}
\label{testbed}
\begin{figure}
    \centering
    \includegraphics[width=1\columnwidth]{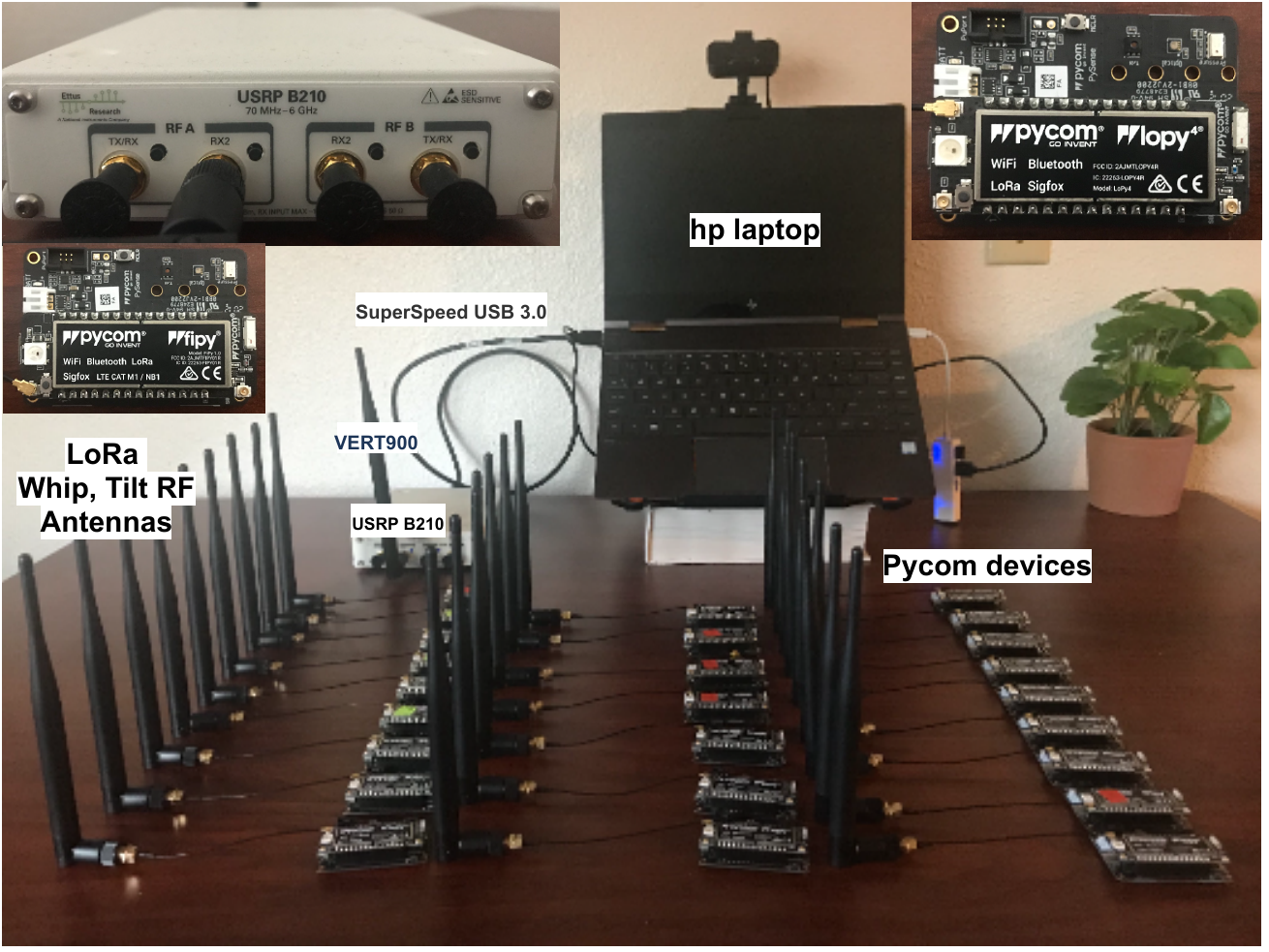}
    \caption{Experimental LoRa device testbed.}
    \label{fig:dataset}
\end{figure}

Fig.~\ref{fig:dataset} depicts our testbed, consisting of $25$ Pycom LoRa-enabled devices and Ettus USRP B$210$ receivers. USRP receivers used for signal sampling were configured at a center frequency of $915$MHz and a sampling rate of $1$MS/s. 
Each device was connected to a dedicated antenna and configured for Raw-LoRa mode transmissions with: $125$KHz bandwidth, a spreading factor (SF) of $7$, a preamble of $8$, a TX power of $20$dBm, and a coding rate of $4/5$. 
We used MicroPython to program and configure the Pycom transmitter boards, and GNURadio software 
to configure the USRP B$210$ receivers to capture and preprocess LoRa transmissions and to store sampled data into files. 
PyCom devices were configured to transmit their data using the LoRa protocol. LoRa uses a tunable parameter, termed spreading factor (SF), to adjust the data rates and energy usage so at to compensate for signal strength degradation or power capacity limitation. SF ranges from 7 to 12, with higher values yielding longer ranges and lower data rates.

\subsection{Experimental Scenarios and RF Datasets}
\label{sec:experiment}
Our RF datasets, collected using the testbed described in Section~\ref{testbed}, provide both time-domain IQ samples and their corresponding FFT representations for multiple different scenarios, specifically designed to enable thorough assessment of fingerprinting approaches.
In all scenarios, we consider a total bandwidth of $1$MHz that covers in-band as well as an adjacent out-of-band spectrum of LoRa transmissions. The reason for choosing this bandwidth is explained in Section~\ref{sec:proposed} when we present our proposed technique.
Recorded IQ and FTT data samples are stored into binary files in compliance with the Signal Metadata Format (SigMF),
where a metadata file written in plaintext JSON (JavaScript object notation) is created for each binary file to include recording information such as sampling rate, recording time/day, carrier frequency, and other parameters.
In this work, we use the following experimental scenarios (The complete RF datasets---with a more comprehensive scenario set, detailed description and downloading links---can found at~\cite{elmaghbub2021comprehensive}).

\myitemizebeginless
\item {\bf Different Days (Indoor and Outdoor) Scenario,} with data collected over multiple consecutive days using the same receiver.
For each day, each device generated ten $20$s-long transmissions, separated one minute apart from each other.
This multi-day scenario allows us to study the robustness of deep learning models when trained on data collected on one day but tested on data captured on a different day.

\item{\bf Different Locations (Indoor and Outdoor) Scenario,} to study of how well the deep learning models perform when trained and tested on data captured in different locations.
For this, we considered three different locations/environments---room, office, and outdoor, and captured LoRa transmissions at each of these locations, all taken on the same day using the same receiver.
One $20$s-long transmission is collected from each device at each location, with $1$-minute duration separating the collections from the consecutive devices.

\item{\bf Different Configurations (Indoor) Scenario,} to study the sensitivity of fingerprinting to LoRa configuration changes.
Ideally, the learning models should still perform well even when tested using an SF configuration that is different from that used for training. To assess this, we collected data samples under:~\\
{\em Configuration 1}: SF = 7; bit rate = 5470 bps~\\
{\em Configuration 2}: SF = 8; bit rate = 3125 bps ~\\
{\em Configuration 3}: SF = 11; bit rate = 537 bps~\\
{\em Configuration 4}: SF = 12; bit rate = 293 bps.~\\
All four configurations have a transmit power of 20dBm, a bandwidth of 125KHz, and a coding rate of 4/5.
For each configuration, one $20$s-long transmission was collected from each device in an indoor setup.
All experiments were taken on the same day using the same receiver.

\item{\bf Different Receivers (Indoor) Scenario,} to assess the sensitivity of the learning models to changes in the receiver hardware. This is because receivers, like transmitters, also have hardware impairments that can impact the robustness of the learning models.
For this, we collected data from the 25 Pycom transmitting devices with two identical USRP B$210$ receivers.
For each receiver, one $20$s-long transmission was captured from each transmitter in an indoor setup.

\myitemizeendless

In all these experiments, in order to allow for the study of the impact of one parameter at time (e.g., time, location, configuration, and receiver), we chose a transmitter-receiver distance that can be fixed for all scenarios. Because $5m$ was the maximum feasible distance in the indoor case, we set that distance to $5m$ for all cases. Unless mentioned otherwise, SF is set to $7$, which is the default value of LoRa protocol.

\section{Transmitter Hardware Impairments and Their Out-of-Band Spectrum Emissions}
\label{sec:hw}
Hardware imperfections originated during manufacturing of the various hardware components of the transmitter, such as the local oscillator (LO), power amplifier (PA), and mixer, cause the characteristics of transmitted RF signals to deviate from their ideal values. This deviation creates unique fingerprints of the transmitting devices that can be exploited to identify them.
Despite the engineering efforts aimed at minimizing these impairments, they cannot be removed completely, leading to some remaining distortions in the received RF signal that can still be exploited to serve as device fingerprints.
For illustration, we use in this paper the phase noise impairment caused by the local oscillator's imperfections to explain how such an impairment can be exploited for device fingerprinting. Detailed analysis of this, as well as other hardware impairments, can be found in~\cite{elmaghbub2020widescan}.

\subsection{Phase Noise Caused by Local Oscillator}
Typical direct conversion transmitters use quadrature mixer configuration to upconvert, at the carrier frequency $w_c$, the in-phase $(I)$ and quadrature $(Q)$ components of the baseband signal using two independent mixers, one used for the $I$ and one for the $Q$ components. First, using the two mixers, the $I$ and $Q$ components are each multiplied (separately and in parallel) with one of the two oscillating signals, one coming from the LO port and the other also coming from the LO port but shifted by 90°.
The outputs of the two mixers are summed together to form the upconverted signal modulated at the carrier frequency $w_c$.
For an ideal LO, the periodic oscillating signal generated by the LO can be represented as a pure sinusoidal waveform $\cos(w_ct)$, which preserves the original spectrum shape in the upconversion operation.
For real scenarios, however, the time domain instability of the oscillating signals generated by real LOs causes random phase fluctuations in the signal, which results in a spectral expansion around the spectrum of the carrier signal. More formally, the oscillating signal generated by real LOs can be expressed as $\cos(w_ct + \theta(t))$, where $\theta(t)$ represents the phase noise, which results in a random rotation of the signal constellation observed at the receiver and thus incurs undesired out-of-band (OOB) emission.

\subsection{Out-of-Band Emissions Caused by Phase Noise}
\begin{figure}
    \centering
    \includegraphics[width=1\columnwidth]{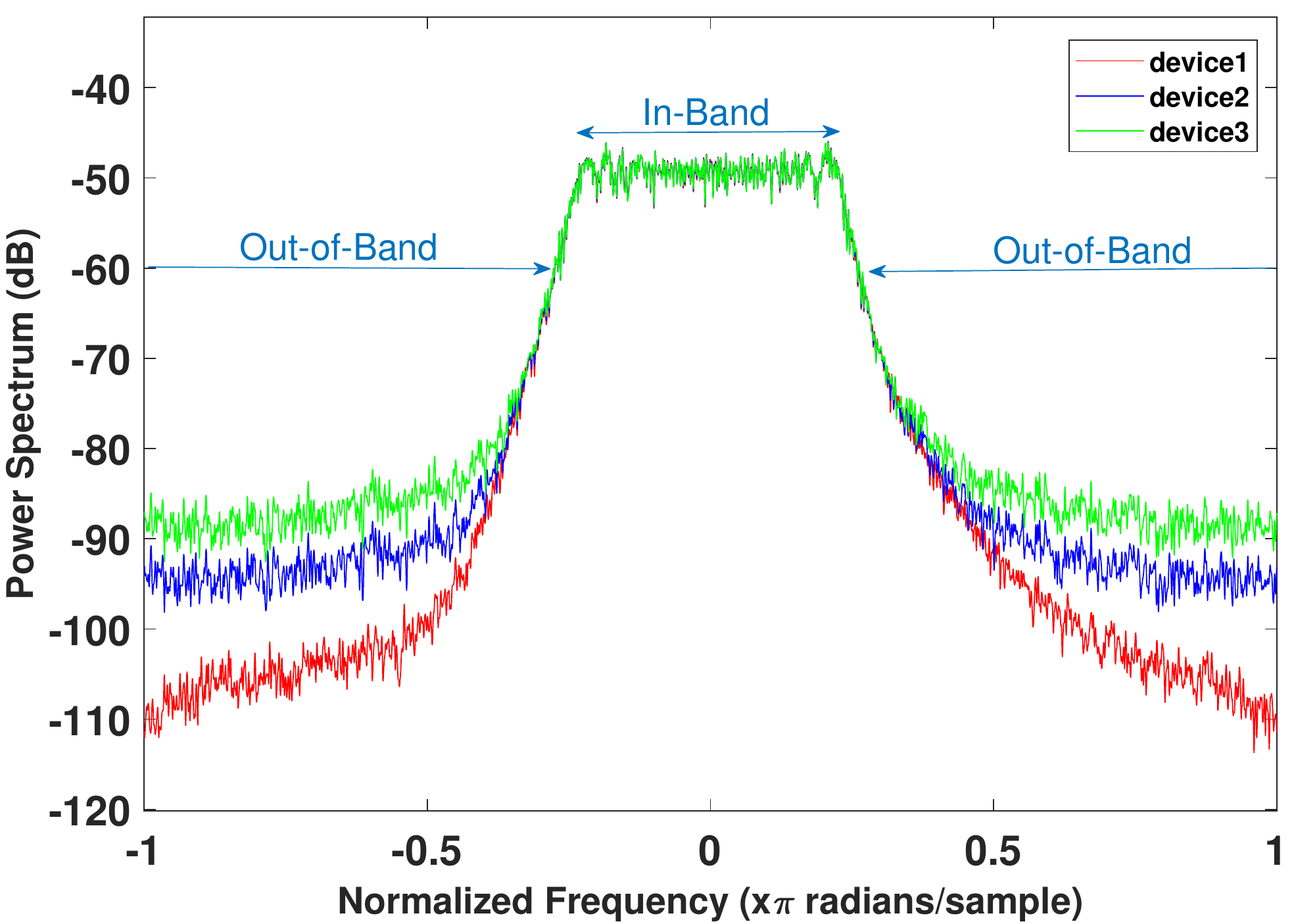}
    \caption{Normalized 1MHz spectra of a LoRa signal (Device 1) and two variants of the same signal distorted with two phase noise magnitudes, 0.2 (Device 2) and 0.4 (Device 3).}
    \label{fig:LO}
\end{figure}

To illustrate this OOB emission, consider applying the Fourier transform to the output of an in-phase mixer, $S_I(t)\cos(w_ct+ \theta(t))$, which is the product of the in-phase baseband signal, $S_I(t)$, and the real LO signal, $\cos(w_ct + \theta(t))$. Straightforward Fourier analysis shows that the phase noise term, $\theta(t)$, results in a bandwidth expansion beyond the original signal’s spectrum around the carrier frequency $w_c$. This bandwidth expansion is originated from the convolution of the spectrum of the upconverted signal centered around $w_c$ and that of the phase noise (or precisely Fourier transform of $e^{–j\theta(t)}$). We refer the readers to~\cite{elmaghbub2020widescan} for formal expressions of this Fourier analysis. The characteristics of LO-originated spectrum regrowth depend on the magnitude of the LO’s phase noise, which varies from one device to another, and therefore, different devices will exhibit different OOB distortions. This can be seen in Fig.~\ref{fig:LO}, in which we plotted the spectra of one LoRa transmission from one of the Pycom devices (Device 1) and two distorted variations of the same signal with phase noise magnitude equaling to $0.2$ (Device 2) and $0.4$ (Device 3). The figure clearly shows that the OOB spectrum shapes of Device 2 and Device 3 are different from one another and from that of Device 1. As shown later, our technique exploits these OOB frequency features generated by the phase noises and other hardware impairments to enhance device fingerprinting accuracy.

\section{Exploiting Out-of-Band Emission Information for Distinctive Device Fingerprints}
\label{sec:proposed}
In the previous section, we showed how hardware impairments cause out-of-band (OOB) emissions in the spectrum surrounding the in-band region that exhibit distinctive device features. In this section, we propose to leverage these distinctive features to enhance device fingerprinting accuracy. The extraction of such features from the surrounding spectrum could be achieved via oversampling or extending the filtered band to include the OOB spectral region. The novelty of the proposed technique lies therefore in the fact that the OOB distortions caused by hardware impairments are unique to the transmitting devices, thereby providing signatures that can be used to uniquely identify them. The proposed fingerprinting technique (i) is resilient against feature cloning and modification since it inherently relies on hardware-driven features that are too difficult to temper with, (ii) does not impose changes on the transmitters, and (iii) incurs little processing at the receiver side that can be done with existing radio technology.

\subsection{Classification Accuracy}
\begin{figure}
    \centering
    \includegraphics[width=\columnwidth]{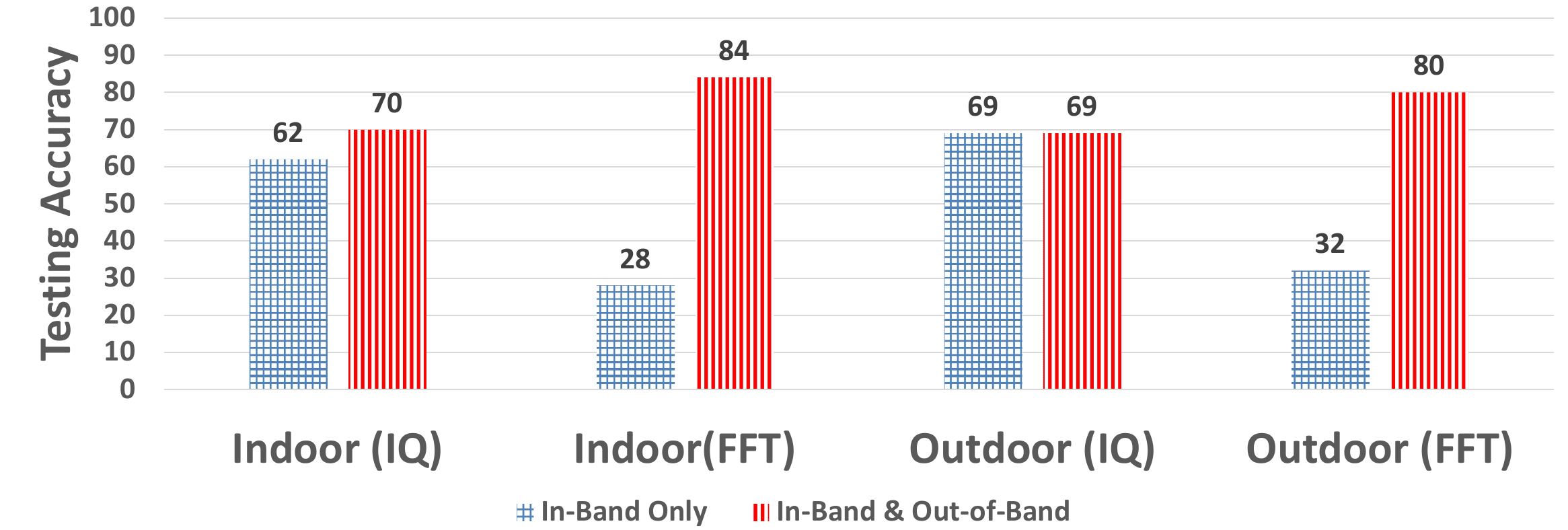}
    \caption{Conventional (in-band only) versus proposed (in-band and OOB) under both Indoor and Outdoor Scenarios.}
    \label{fig:eval0}
\end{figure}

To assess the effectiveness of the proposed technique, we implemented a Convolutional Neural Network (CNN) model (see Section~\ref{subsec:CNN-arch-model} for detail) and evaluated the classification accuracy, obtained under each of the proposed (in-band and OOB) and conventional (in-band only) techniques, using our RF datasets described in Section~\ref{sec:experiment}.
Fig.~\ref{fig:eval0}~shows the average testing accuracy achieved under both techniques for the Indoor and Outdoor setups while considering both domain-time IQ and frequency-domain FFT representations as an input to the learning models.
First, note that when the FFT is used as the input to the learning models, the proposed (in-band and OOB) approach achieves a much higher accuracy compared to the conventional approach. More specifically, in both indoor and outdoor scenarios, the proposed approach achieves an accuracy of about $80\%$ whereas that achieved under the conventional one is only about $30\%$. This increase in the accuracy is due to the fact that the proposed technique extracts its features from both the signal's in-band spectrum and the adjacent spectrum, thereby capturing out-of-band distortion information caused by the hardware impairments as explained in Section~\ref{sec:hw}. Second, observe that the proposed approach does slightly better than the conventional approach only in the indoor setting when IQ is used as the input to the learning models. More detail on the evaluation setup and learning model architecture is provided in Section~\ref{sec:Evaluation}.

\subsection{Computation Cost}
System complexity and training time are other essential metrics when proposing authentication techniques as they impact the speed of the process and challenge the scalability. Our proposed technique does not incur additional computational costs at the transmitters, and instead, it only requires a slightly higher sampling rate at the receiver side, which is feasible without a considerable overhead due to the advances in the sampling capabilities of radio technologies. Furthermore, compared to the in-band only approach, our technique does not result in an increase in the training time, nor does involve more model parameters.

\section{Revealing LoRa Device Fingerprinting Sensitivity: Experimental Results}
\label{sec:Evaluation}
\begin{figure*}
    \centering
    \includegraphics[width=2\columnwidth]{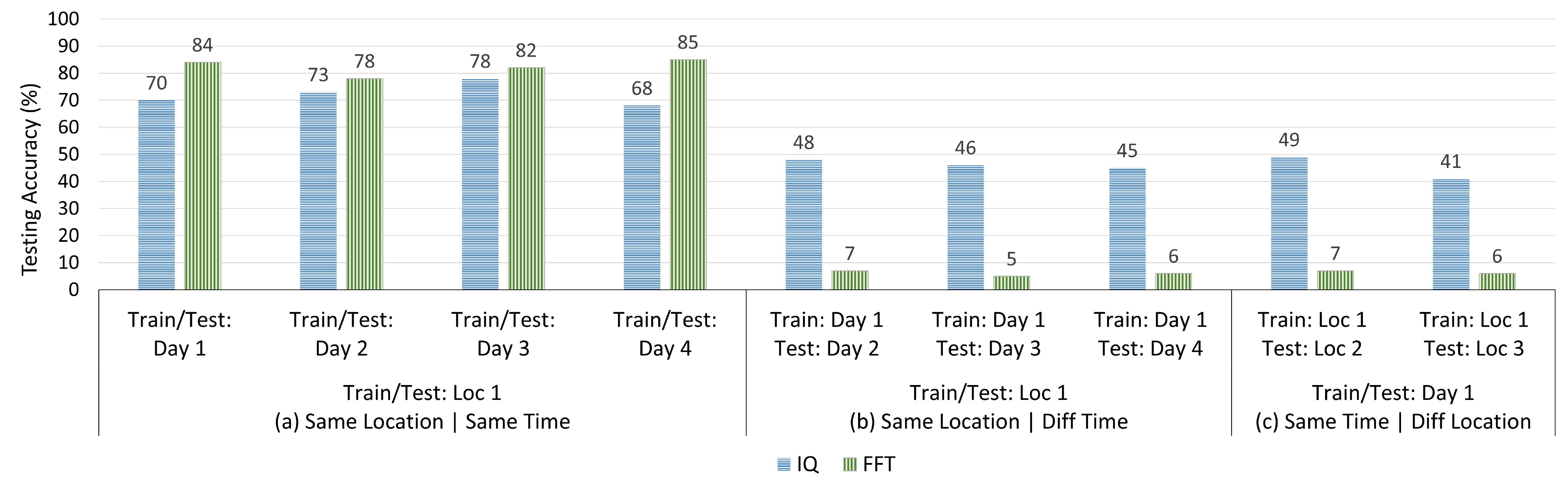}
    \caption{Sensitivity to changes in channel conditions due to temporal and spatial variability. USRP Receiver 1 is used for data sampling and collection, and LoRa protocol configuration is set to Config 1 (SF=7). Loc 1 is an indoor/room environment, Loc 2 is an outdoor environment, and Loc 3 is indoor/office environment.}
    \label{fig:channel-variability}
\end{figure*}

In this section, we present experimental results to disclose and understand the performance sensitivity of LoRa device fingerprinting and classification to network deployment and channel condition variability. Throughout this section, we only study the proposed technique that we discussed in Section~\ref{sec:proposed}.

\subsection{CNN Architecture and Parameters}\label{subsec:CNN-arch-model}

Deep CNNs have been the most commonly used architectures in deep-learning-based RF fingerprinting with the state of the art performances. And hence, CNNs have been perceived as the appropriate candidate for showing the advantage of our technique. For our evaluation, we employ a CNN model similar to the one used in~\cite{liu2017deep}, where
the data input, represented as 2-dimensional (one for real, one for imaginary) tensor, is fed to the first convolution layer of the CNN. This first layer has sixteen $1$x$4$ filters, each learning I and Q time variations, all generating $16$ distinct feature maps (with the exception of the last convolution layer whose filters are of size $2$x$4$ generating features for both I and Q at the same time).
After each (except the last) convolution layer there is a Batch Normalization layer, a Leaky Rectified Linear Unit activation, and a Maximum Pooling layer of [$1$ $2$] stride and $1$x$2$ size.
The last convolution layer is followed by a $1$x$256$ Average Pooling layer whose output is fed an input to $25$-neuron Fully Connected layer, which is then followed by another activation layer and dropout layer (dropout rate = $0.5$).  The output of the Fully Connected layer is fed at last to the classifier layer. The probabilities of each input frame is produced by a Softmax classifier employed at the last layer of the CNN.

Stochastic Gradient Descent is used to update the CNN weights. The momentum optimizer of the gradient descent has an initial learning rate, rate drop factor, rate drop period, and a L2-regularization value of $0.07$, $0.1$, $19$, and $0.0001$, respectively. The categorical cross-entropy is used as the loss function and the back-propagation method is used to minimize the prediction error.
We used MATLAB's Deep Learning Toolbox to implement the CNN model, and run the classifier using our collected datasets (with $80\%$ for training, $10\%$ for validation, and $10\%$ for testing) on a computing cluster containing 6 Nvidia dgx2 nodes with 16 V100 GPUs. For CNN's input data, we used a sliding window of size $8192$ samples with stride size that is equal to the window size.

\subsection{Input Data Representation}\label{subsec:data-rep}
We study the use of two different representations of the data input to the learning model: raw time-domain IQ data and its frequency-domain FFT representation.

\subsubsection{IQ Representation}
IQ representation frames are generated by first creating fixed-size 1D complex-valued vectors, whose real and imaginary parts correspond to the I (in-phase) and Q (quadrature) components, and then converting these 1D complex-valued vectors into two 1D real-valued vectors, $IQ$, with each dimension carrying the I and Q samples.

\subsubsection{FFT Representation}
This provides the spectral components of the sampled signal to reveal its frequency features, thus capturing the spectrum artifacts generated by the various hardware impairments. FFT representation frames are generated by first converting the time-domain signal samples into frequency domain using MATLAB's {\tt fft} function. Then the resulted 1D complex-valued vectors is converted into two 1D real-valued vectors, one carrying the real part and one carrying the imaginary part.

\begin{figure*}
    \centering
    \includegraphics[width=2\columnwidth]{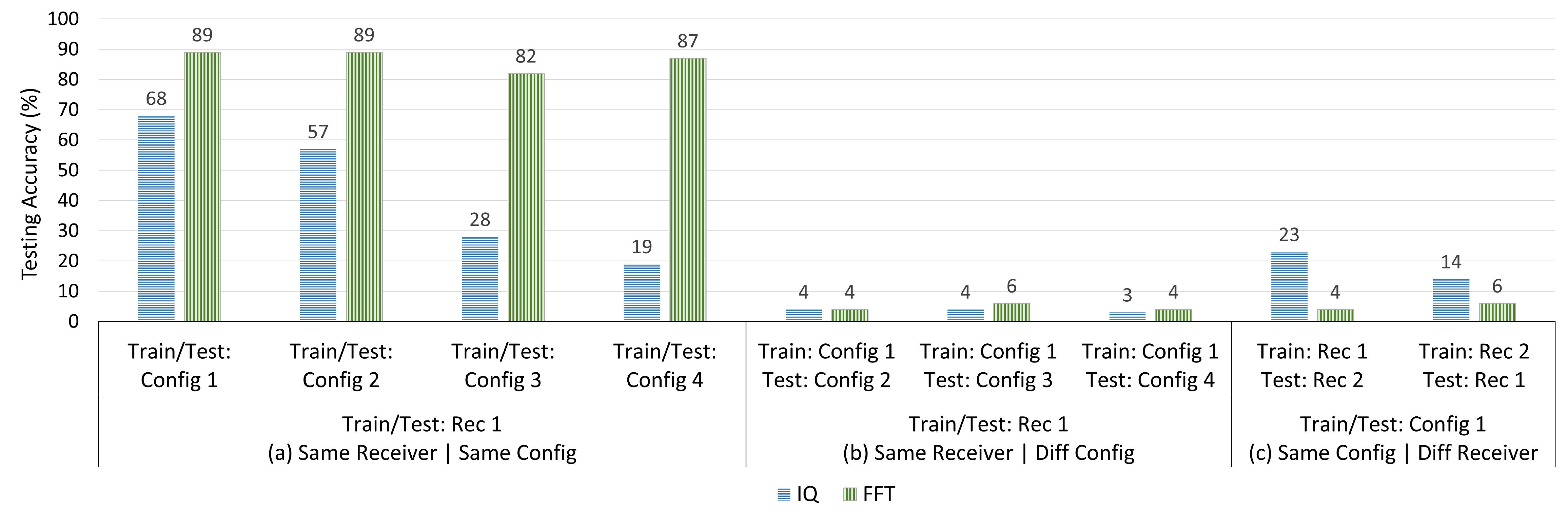}
    \caption{Sensitivity to changes in LoRa SF configuration and USRP receiver hardware. All experiments are conducted during the same day at the same location, Loc 1 (indoor/room).}
    \label{fig:system-variability}
\end{figure*}

\subsection{Sensitivity of Fingerprinting Accuracy to Wireless Channel Condition Variability}
Fig.~\ref{fig:channel-variability} shows the fingerprinting accuracy behavior observed when the deep learning models are trained and tested under same/different channel conditions while considering both IQ and FFT representations as input to the models.
Fig.~\ref{fig:channel-variability}(a) reveals the accuracy when training and testing are done using the same channel conditions; i.e., the training and testing data is collected during the same time at the same location, Loc 1, which is an Indoor (room) environment.
Figs.~\ref{fig:channel-variability}(b) and~\ref{fig:channel-variability}(c) show the accuracy achieved when the channel conditions used during training are different from those used during testing, due to temporal (Fig.~\ref{fig:channel-variability}(b)) or spacial (Fig.~\ref{fig:channel-variability}(c)) variations. The results shown in Fig.~\ref{fig:channel-variability}(b) are based on the Different Days Scenario (Indoor) datasets, and those shown in Fig.~\ref{fig:channel-variability}(c) are based on the Different Locations Scenario datasets, all described in Section~\ref{sec:experiment}. Similar results were obtained for the Outdoor scenarios, but omitted here due to space limitation.

We begin by looking at the results depicted in Fig.~\ref{fig:channel-variability}(a), when the channel conditions used for training and testing are the same.
First, observe that the achieved accuracy is consistent throughout each of the 4 studied days.
Second, observe that the FFT data representation achieves an accuracy of about $10\%$ higher than that achieved under the IQ representation.

Now when the training and testing are done under different channel conditions, whether due to time change (Fig.~\ref{fig:channel-variability}(b)) or due to location change (Fig.~\ref{fig:channel-variability}(c)), the accuracy drops substantially for both types of input representation, IQ and FFT.
Our findings are consistent with previous studies, indicating that the learning models tend to latch into channel-related as opposed to device-related features, making them unable to maintain their high accuracy under channel condition changes.

Another interesting observation we make is that when training and testing are performed under different channels, FFT performs worse than IQ, regardless of whether the channel condition variation occurs due to a temporal or a spatial change. In both types of change, the accuracy drops from about $72\%$ to about $46\%$ under IQ but from about $82\%$ to about $6\%$ under FFT.
Our conclusion is that the learning models seem to be more resilient to channel condition changes when using time-domain IQ representation as its input as opposed to using the frequency-domain FFT representation. This suggests that the impact of channel impairment variations is more profound in the frequency domain than in the time domain.

\subsection{Sensitivity of Fingerprinting Accuracy to System Setting Variability}
\begin{figure}
    \centering
    \includegraphics[width=1\columnwidth]{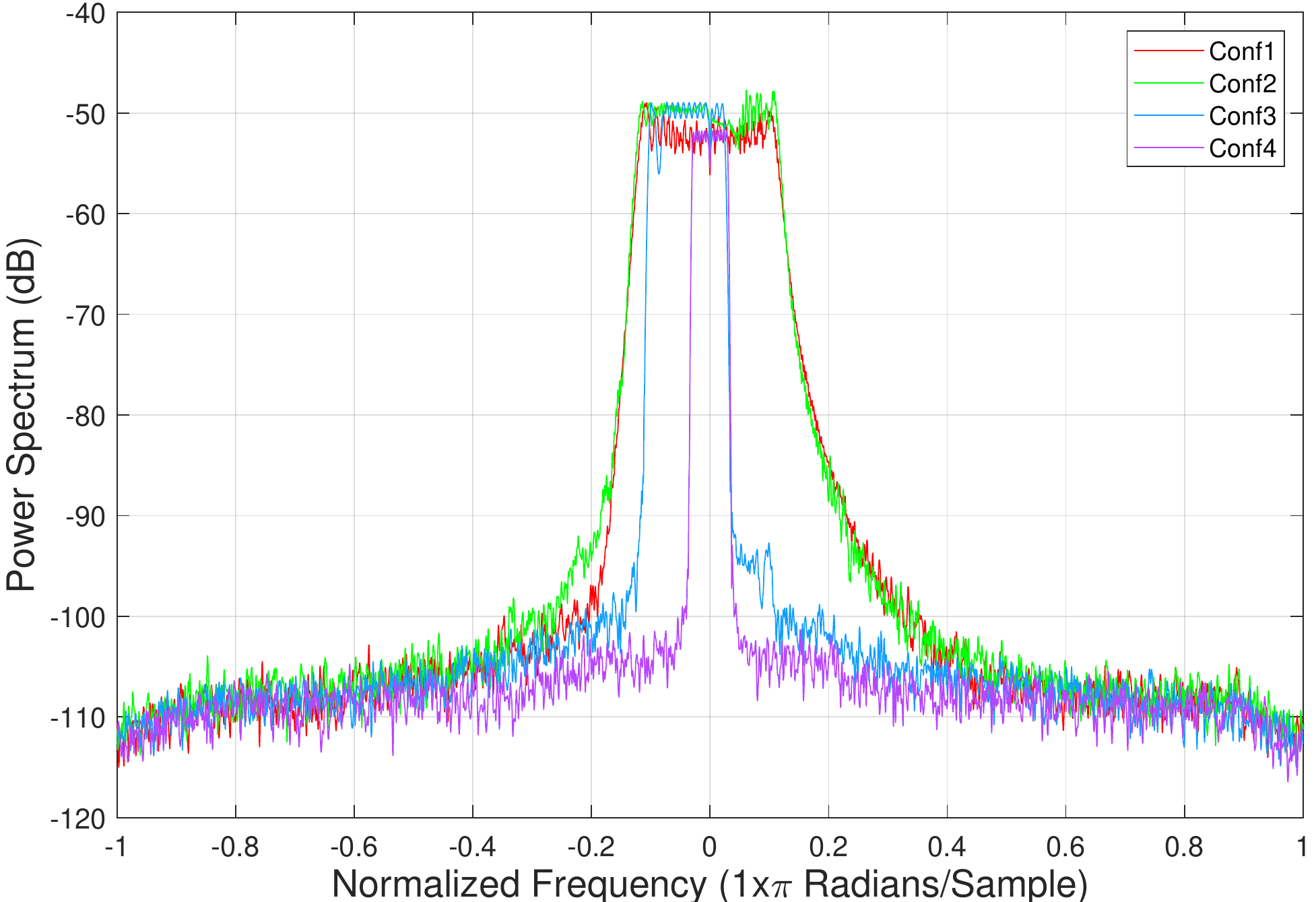}
    \caption{Spectrum visualization of 4 LoRa configurations: Conf1, Conf2, Conf3 and Conf4 correspond to SF=7, SF=8, SF=11 and SF=12, respectively.}
    \label{fig:spec4}
\end{figure}

Fig.~\ref{fig:system-variability} shows the accuracy when the training and testing are performed while keeping the same wireless channel conditions (same time and same location) but varying the LoRa protocol configuration and/or the receiver hardware. Specifically, it presents the fingerprinting accuracy behavior when training and testing are done under: same LoRa configuration and same receiver hardware (Fig.~\ref{fig:system-variability}(a)), different LoRa configuration but same receiver hardware (Fig.~\ref{fig:system-variability}(b)), and same LoRa configuration but different receiver hardware (Fig.~\ref{fig:system-variability}(c)). For this experiment, we consider the four different LoRa configurations described in Section~\ref{sec:experiment} and two identical USRP B$210$ receivers: Rec 1 and Rec 2. Recall that, as discussed in Section~\ref{sec:experiment}, changing the spreading factor (SF) of LoRa modulation changes the data rate, reception sensitivity, and power consumption, which in turn results in a change of the shape of the spectrum. Fig.~\ref{fig:spec4} shows spectrum shapes of the four LoRa configurations considered in our datasets, which are described in Section~\ref{sec:experiment}. Note the difference in the spectrum shape observed under each of the four LoRa configurations.

From Fig.~\ref{fig:system-variability}(a), we first observe that FFT input representation consistently outperforms IQ representation when same configuration is used during training and testing. Also, we observe that for the IQ case, the higher the spreading factor, the higher the degradation in the accuracy.
Now surprisingly, as can clearly be seen from Fig.~\ref{fig:system-variability}(b), the learning models perform very poorly, under both IQ and FFT input representations, when tested and trained using different SF configurations, even when data capturing is done during the same time and at the same location. The figure shows that the testing accuracy decreases significantly to about $4\%$ (what random guesses achieve) when configuration 1 is used for training while configuration 2, 3 or 4 is used for testing, and this is regardless of whether FFT or IQ is used for data input.

Fig.~\ref{fig:system-variability}(c), showing the accuracy behavior when using a receiver during testing that is different from that used during training, allows us to assess the impact of the receiver's impairments on the performance. This is done by using two identical USRP B$210$ receivers (Rec 1 and Rec 2) that capture the transmissions of the $25$ devices at the same location (indoor/room), using the same LoRa configuration, and during the same time. This set of results is obtained based on the RF datasets presented in the Different Receivers Scenario, explained in Section~\ref{sec:experiment}.
Like in the case of changes occurring in the LoRa configuration, Fig.~\ref{fig:system-variability}(c) shows that the accuracy drops significantly when training and testing are done on different receivers, with a lesser severity in the case of IQ representation; this is true for both: train on Rec1 / test on Rec2 and train on Rec2 / test on Rec1.

We conclude that unlike when changes occur in the channel conditions, the learning models completely lose their ability to classify devices when changes occur in the LoRa configuration and/or the receiver hardware, implying that changes in the system setting yield drastic changes in the data distribution, severely crippling fingerprinting performances.

\section{Open Issues and Future Directions}
\label{open-challenges}
We next highlight a few key open challenges that require further investigation:

\begin{itemize}
\item We view the randomness and unpredictability of the wireless channel as one of the main obstacles that hinder the real-world deployment of data-driven RF fingerprinting techniques. 
Deep-learning-based RF fingerprinting approaches are still not robust to distribution shifts due to changes in the network deployment environment. The real-world applicability of these techniques is at stake as long as the robustness issue is not fundamentally solved.

\item We believe that gaining some transparency into what these deep learning-based models actually learn and then being able to encourage them to latch onto the hardware-specific (the real signature), not time-variant nor channel-related RF features, would significantly increase the robustness of existing deep learning models.

\item So far, most research efforts have focused on fingerprinting and classifying known devices in a closed set scenario. Shifting the RF fingerprinting problem into the 'open set' classification realm would be crucial in promoting and accelerating the adoption of these techniques in the real world. Semi-Supervised learning and meta-learning have the potential to play a critical role in this transformation.

\item Lastly, the overall security of machine learning models is another vital issue that has gained a lot of interest recently. Adversarial poisoning attacks, for example, can impact the prediction functionality of deep-learning-based techniques. Therefore, the robustness of deep-learning-based RF fingerprinting models against Adversarial and other attacks should be given special attention in the near future \cite{lalouani2021countering}.
\end{itemize}

\section{Conclusion}
\label{sec:Conclusion}
This paper presents an experimental framework that investigates the robustness of deep-learning-based RF fingerprinting of LoRa-enabled devices against changes in the network deployment environment. We propose a new fingerprinting technique that exploits out-of-band distortion information caused by hardware impairments to provide device-specific signatures that can be used to enhance fingerprinting performances. Using LoRa RF datasets, we show that the deep learning models perform relatively well when trained and tested under the same network deployment settings, but poorly when trained and tested under different settings.

\section{Acknowledgement}
This research is supported in part by the National Science Foundation and Intel under Award No. 2003273.  
We would like to thank Intel researchers, Dr. Kathiravetpillai Sivanesan, Dr. Lily Yang, Dr. Richard Dorrance, and Dr. Vesh Raj Sharma Banjade, for their constructive feedback. 

\label{sec:Acknow}

\bibliographystyle{IEEEtran}

\bibliography{References}

\begin{IEEEbiography}[{\includegraphics[width=1in,height=1in,clip,keepaspectratio]{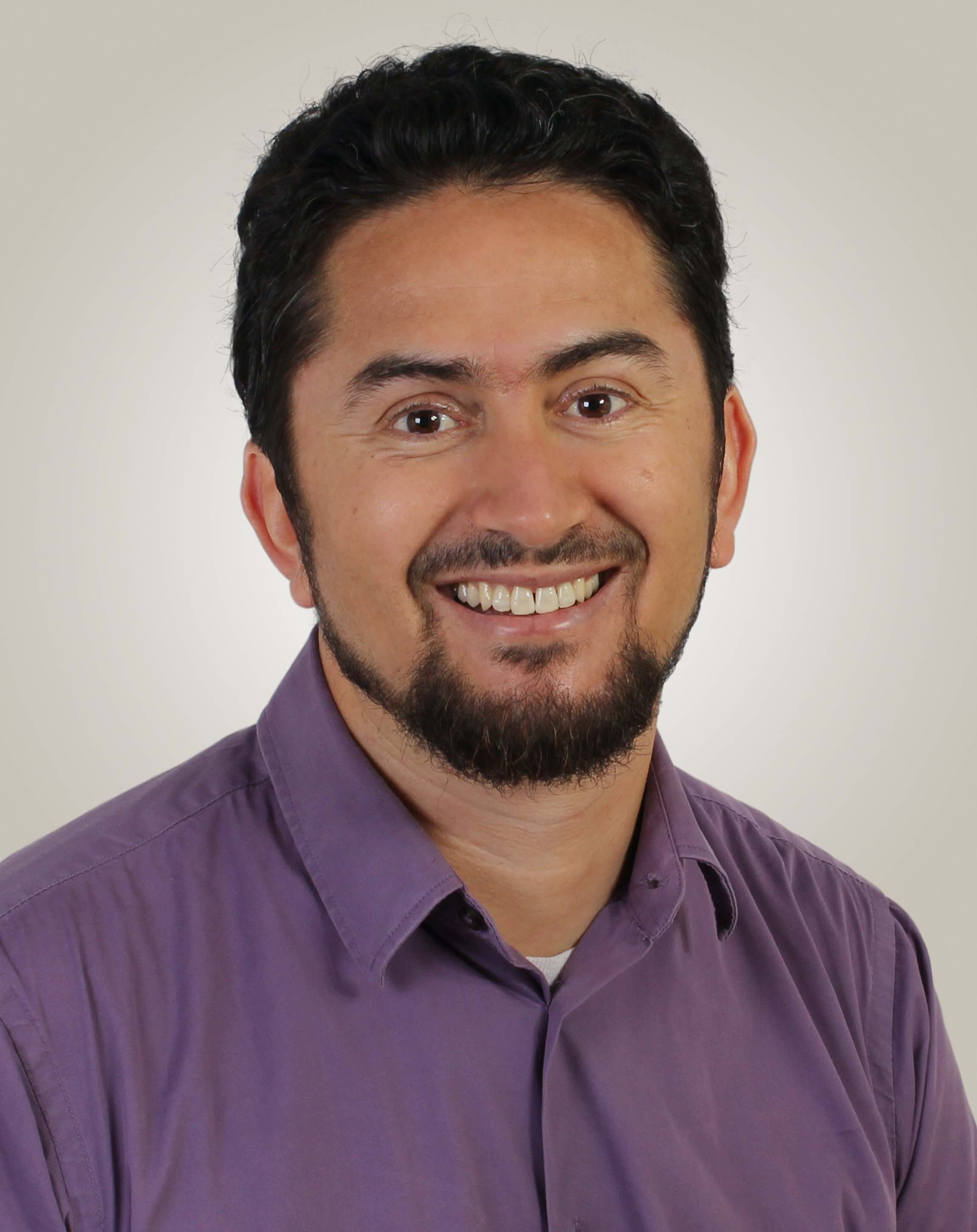}}]{Bechir Hamdaoui} is a professor in the School of EECS at Oregon State University. His research interests are in the general areas of networked systems, wireless \& network security, and computer networks. He currently serves as the Chair for the IEEE Communication Society (ComSoc)'s Wireless Communications Technical Committee (WTC).
\end{IEEEbiography}

\begin{IEEEbiography}[{\includegraphics[width=1in,height=1.25in,clip,keepaspectratio]{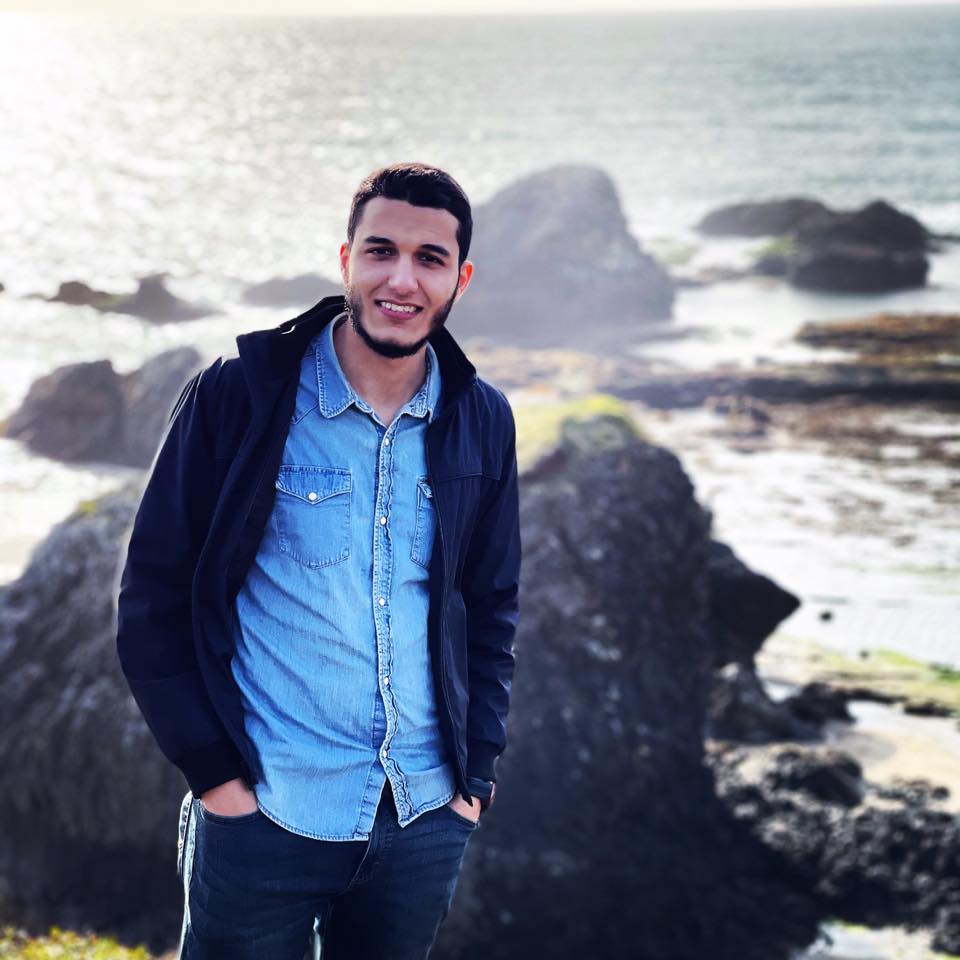}}]{Abdurrahman Elmaghbub} received the B.S. (2019) and M.S. (2021) degrees in ECE from Oregon State University, where he is currently pursuing his Ph.D. degree. His research interests include wireless communication and networking, with a current focus on applying deep learning to wireless device classification.
\end{IEEEbiography}

\end{document}